\documentclass[onecloumn]{article}

\usepackage{hyperref} 

\pdfoutput=1

\usepackage[latin1]{inputenc}
\usepackage{mystyle}
\usepackage{url}
\usepackage{fullpage}
\usepackage{cite}
\usepackage{caption}
\allowdisplaybreaks
\usepackage{amsmath,graphicx}
\usepackage{authblk}

\graphicspath{{./images/}}

\hypersetup{  
  bookmarks=true,
  backref=true,
  pagebackref=false,
  colorlinks=true,
  linkcolor=blue,
  citecolor=red,
  urlcolor=blue,
  pdftitle={EOT-GAN},
  pdfauthor={Dong Liu},
  pdfsubject={}
}

\title{Entropy-regularized Optimal Transport Generative Models}
\author{
  Dong Liu,
  Minh Thành Vu,
  Saikat Chatterjee,
  and Lars~K. Rasmussen
}

\affil{
  KTH Royal Institute of Technology, Stockholm, Sweden, email: \{doli, mtvu, sach, lkra\}@kth.se}

\begin{document}
\date{}
\maketitle

\begin{abstract}
  We investigate the use of entropy-regularized optimal transport (EOT) cost in developing generative models to learn implicit distributions.
  Two generative models are proposed. One uses EOT cost directly in an
  one-shot optimization problem and the other uses EOT cost iteratively in an adversarial game. 
  The proposed generative models show improved performance over contemporary models for image generation on MNSIT.
\end{abstract}
{\bf Keywords:}
Optimal transport, generative models. 

\section{Introduction}

Data-driven learning of a probability distribution by a generative model is an important problem 
in statistical signal processing and machine learning. 
Recently neural network based generative models are popular tools to
study underlying probability distribution of datasets. A prominent example is 
generative adversarial network (GAN) \cite{NIPS2014_5423}, which learns 
implicit distribution models. 

In the GAN of \cite{NIPS2014_5423}, a generator produces
synthetic samples and a discriminator endeavors to
distinguish between real samples and synthetic samples. Generators and discriminators are realized using (deep) neural networks.
Discriminator and generator
play an adversary game against each other using a `min-max' optimization to learn parameters of neural networks. 
For generator, the game
turns out be minimizing Jensen-Shannon divergence (JSD) between
target distribution and induced distribution by generator when discriminator is
optimal. Using the same adversary
optimization, deep convolutional neural network based GAN (DCGAN)
\cite{2015arXiv151106434R} producing good quality synthetic images, has attracted high attention.

JSD has limitations in GANs where generators and dis-
criminators are based on deep neural networks. The first limitation is that back
propagation suffers from vanishing gradient. Gradient of cost
function with respect to (w.r.t.) generator vanishes as
discriminator approaches optimal (see Theorem~2.4~\cite{2017arXiv170104862A}),
which stops generator from further learning. The second limitation is due to high sensitivity of JSD to slight perturbations. JSD can be large between a distribution $P_x$ and a distribution $P_{x+\epsilon}$ where $\epsilon$ is perturbation\cite{2017arXiv170104862A}. 

Both limitations are addressed in Wasserstein GAN (WGAN)
\cite{2017arXiv170107875A}. Wasserstein distance stems from
optimal transport (OT) problem, which measures divergence between two
distributions. The WGAN formulation does not require an explicit
discriminator and it does not has the vanishing-gradient problem. Further,
Wasserstein distance/OT is upper
bounded by the standard deviation of perturbation $\epsilon$
\cite{2017arXiv170104862A}, addressing the second limitation. 

OT based cost in WGAN brings a strict constraint to follow in its optimization.
Kantorovich-Rubinstein duality used in WGAN requires a supremum over
infinite set of all Lipschitz functions with Lipschitz constant equal
to one. Various sub-optimal techniques are proposed to enforce the
Lipschitz property. An example is weight clipping
\cite{2017arXiv170107875A} where neural network parameters (weights)
are updated first without Lipschitz constraint and then projected to
satisfy Lipschitz constraint in each iteration. Other approaches are
gradient penalty\cite{2017arXiv170400028G} and spectrum normalization\cite{2018arXiv180205957M}.

In this article, our main contribution is to explore use of Entropy-regularized OT (EOT) cost for generative models. 
The EOT was studied earlier for efficient comparison between two probability distributions \cite{2013arXiv1306.0895C}. 
The major advantage of EOT is
that corresponding dual problem is free from Lipschitz constraint. Use of EOT improves
analytical tractability allows us to develop two generative models. Our
first model considers EOT cost directly on distribution of signals (in our case,
on the image pixels). This model uses an one-shot optimization problem, \emph{i.e.} no use of adversarial game in iterations. The second model
considers EOT on feature distribution instead of direct signal
distribution. In this case we also need to learn a representation
mapping, which is implemented as a neural network. 
This requires alternative optimization of representation mapping and
generator. In addition to the above advantage, duality of EOT can be
effectively solved and straightforwardly extended to parallel computation.

\section{EOT based generative models}

In this section, we begin with Entropy-regularized OT (EOT) cost and then propose generative models.

\subsection{Entropy-regularized OT} 

We denote our working space by $(\mathcal{X},\|\cdot\|_2)$ where
$\mathcal{X}\subset\mathbb{R}^d$ and $\|\cdot\|_2$ is the Euclidean
distance. Assume that $\mathcal{X}_1$, $\mathcal{X}_2$ are $N$-sample
subsets of $\mathcal{X}$. Let $P$ be a distribution on $\mathcal{X}_1$
and $Q$ be a distribution on $\mathcal{X}_2$. OT calculates the
minimum cost of transporting distribution $P$ to $Q$. We use $W(P,Q)$ to denote entropy-regularized OT (EOT) cost as follows:
\begin{equation}\label{eq-entropic-wsd}
  W(P,Q)=\min_{ \pi \in \Pi(P,Q)} \dotp{\pi}{M} - \la H(\pi),
\end{equation}
where $\dotp{\cdot}{\cdot}$ stands for the inner product of
two matrices, and $\Pi(P,Q)$ is a set of joint distribution $\pi$ on
the sample sets $\mathcal{X}_1\times\mathcal{X}_2$ such that $\pi$ has
marginal distributions $P$ and $Q$. 
The cost matrix $M$ has elements $[M]_{i,j} = d(x^{(i)}, y^{(j)}) = \normd{x^{(i)} - y^{(j)}}^2$ and $x^{(i)}, y^{(j)}$ are
samples of $P, Q$, respectively. Here $H(\pi) = \sum_{i,j} -\pi_{i,j}
\log(\pi_{i,j})$ and $\la \in \RR^{+}$ is the regularization
parameter. The entropy regularization in \eqref{eq-entropic-wsd}  translates 
to a requirement that the joint distribution $\pi$ has a high entropy. 
Note that $\|\cdot\|_2$ is invariant of unitary transform and hence
representation of $\mathcal{X}$ in another unitary basis does not
change the cost matrix. The duality of EOT cost in \eqref{eq-entropic-wsd} is
\begin{equation}\label{eq-dual-wsd}
  W(P,Q) \! =  \max_{\al, \be \in \mathbb{R}^{N}} \al^{T}P \! + \! \beta^{T}Q \! - \!
  \sum_{i,j} \lambda e^{ \frac{{\left( \al + \beta - [M]_{i,j} \right)}}{\la} },
\end{equation}
where $\alpha,\beta$ are dual variables and $(\cdot)^T$ means transpose. The optimal dual vector $\beta^{\ast}$
of \eqref{eq-dual-wsd} is a subgradient of $W(P,Q)$ with respect to $Q$.
There exists a computationally efficient algorithm called Sinkhorn
algorithm\cite{2013arXiv1306.0895C, 2013arXiv1310.4375C} to
solve~\eqref{eq-dual-wsd}, which alternatively scales the rows and columns of $M$ matrix. This alternative computation gives a pair of
vectors $(u, v) \in \RR^N_{+} \times \RR^N_{+}$ that defines the optimal primary and dual variavles (see proposition 2 in \cite{2013arXiv1310.4375C}): 
\begin{equation}\label{eq-dual-opt}
  \hspace{-1pt}\pi^{\ast}\hspace{-5pt} =\hspace{-3pt}
  \mathrm{diag}(u) \,\hspace{-1pt} e^{\frac{-M}{\la}} \, \hspace{-4pt}\mathrm{diag}(v),  \beta^{\ast}\hspace{-5pt} =\hspace{-3pt}\frac{\log(u^T)\mathds{1}_N}{N\la} \mathds{1}_N -\frac{\log(u)}{\la}.
\end{equation}
where $\mathrm{diag}(u)$ is a matrix with diagonal entries from vector $u$ and $\mathds{1}_N$ is a column vector with ones.

\subsection{EOT based Generative Models}

In this subsection, we propose two generative models.
We first develop an EOT based generative model handling signals/data directly. This models is referred to as EOT generative model (EOTGM). 
In our second model, we use a
representation mapping where EOT cost is used to optimize the
generative model and representation mapping jointly. The second model is referred as EOT based GAN (EOTGAN).


\subsubsection{EOT based generative model (EOTGM)}\label{subsec-gmeot}
Assume that $P$ is the unknown true probability distribution for a given dataset and $Q$ is
a probability distribution induced by generator $g: \Zz \rightarrow
\Xx$. The generator function $g$ depending on parameter $\theta$,
transforms an input $Z$ from latent space $\Zz$ to signal space
$\Xx$, given $Z \sim P_{Z}$ (usually a Gaussian noise). The generator $g(Z)$ 
induces $Q$. Learning of $Q$ is equivalent to minimization of $W(P, Q)$ w.r.t. generator $g$: 
\begin{equation}\label{eq-entropic-model}
  \underset{g:\mathcal{Z}\to \Xx}{\argmin}\, W(P, Q) = \underset{{\theta}}{\argmin}\, W(P, Q).
\end{equation}
Since $\beta^{\ast}$ in
\eqref{eq-dual-opt} is subgradient of $W(P, Q)$ w.r.t. $Q$, we are able to
optimize the generator $g$ such that the induced distribution $Q$
approximates $P$, using gradient chain rule:
\begin{equation}
  \nabla_{\theta}W(Q, P) = \left(\nabla_{\theta}Q\right)^{T}
  \beta^{\ast},
\end{equation}
Alternatively the optimization problem \eqref{eq-entropic-model} can be addressed by solving $\argmin_{g}\dotp{\pi^{\ast}}{M} $ iteratively using auto-gradient functions in PyTorch\cite{pytorch} or
TensorFlow\cite{tensorflow}, where $\pi^{\ast}$ is primary optimal
variable to \eqref{eq-entropic-wsd} given by \eqref{eq-dual-opt}. We
propose Algorithm~\ref{algo-E-WL} to learn distribution $P$
via minimizing the EOT loss w.r.t parameter $\th$
of generator function $g$.
\begin{algorithm}
  \caption{EOT based generative model (EOTGM)}\label{algo-E-WL}
  \begin{algorithmic}[1]
    \Require{$l$: the update rate at each iteration, $N$: the batch
      size, and $\th_{0}$: the initial parameter for $g$.}
    \While{$\theta$ has not converged}
    \State Sample $\left\{ x^{(i)} \right\}_{i=1}^{N} \sim P$, a batch
    from a real dataset. 
    \State Sample $\left\{ z^{(i)} \right\}_{i=1}^{N} \sim P_{z}$, a batch of noise samples.
    \State Get $\left\{ y^{(i)} \right\}_{i=1}^{N}$ by passing
    $\left\{ z^{(i)} \right\}_{i=1}^{N}$ through $g$.
    \State Calculate the cost matrix $M$.
    \State $\pi^{\ast}$, $\beta^{\ast} \gets$ primary and dual
    solutions of $W(\left\{ x^{(i)} \right\}_{i=1}^{N}, \left\{
      {y}^{(i)}\right\}_{i=1}^{N})$ according Equation~\eqref{eq-dual-opt}.
    \State $\theta \gets \theta - l \left(\nabla_{\th}{Q}\right)^T
    \beta^{\ast}$. (Or back propagate using loss $\dotp{\pi^{\ast}}{M}$)
    \EndWhile
  \end{algorithmic}
\end{algorithm}
\subsubsection{EOT based GAN (EOTGAN)}

In this subsection, we consider representation learning (feature
learning) with which usage of EOT is more meaningful than that
directly in signal space.
It is well-known that Euclidean distance is not well suited to compare
two multimedia signals. For example, Euclidean distance between an
image and its rotated version can be large , but they are visually same. In Algorithm \ref{algo-E-WL} we construct cost matrix $M$ in EOT using 
Euclidean distance between real signals and generated signals. Our new proposal is to transform signal through a representation mapping 
$f: \Xx \rightarrow \Mm$, $\mathcal{M}\subset\mathbb{R}^{m}$ and we compare features in the representation space via EOT. We assume that Euclidean distance
between features in the representation space is more semantically
meaningful. An element of the cost matrix $M_f$ in representation domain (feature domain) is as follows:
\begin{equation}\label{def-similarity}
  d_f(x, y) = \|f(x)-f(y)\|_{2}^{2}.
\end{equation}

Our new objective is joint learning of generator $g$ and representation $f$. A natural question is how to construct $f$ function?
Let $x$ and $\tilde{x}$ are two samples from distribution $P$ and $y$ is a generated signal from distribution $Q$. An intuition is that the representation function $f$ should have the following algebraic property: $ d_f(x, \tilde{x}) + \gamma \leq d_f(x, y) $ for $\gamma > 0$. Construction of $f$ function for all triplets $(x, \tilde{x}, y)$ satisfying the algebraic property is non-trivial.

We take an alternate route using EOT cost. Let us denote that the distribution of $f(x)$ and $f(y)$ by $P_f$ and $Q_f$, respectively. Let $M_f$ be the cost matrix in representation domain and its elements $[M_f]_{i,j} = d_f(x^{(i)}, y^{(j)})$, $x^{(i)} \sim P, y^{(j)}\sim Q$.  Then we learn $f$ and $g$ using alternative optimization, as follows. 
\begin{enumerate}
\item Learning of representation $f$ is minimizing EOT cost
  \begin{equation}\label{eq-sim-in}
    W(P_f, P_f) = \min_{\widetilde{\pi} \in \Pi(P_f, P_f) } \dotp{\widetilde{\pi}}{\widetilde{M}_f} - \la H(\widetilde{\pi}),
  \end{equation}
  where $[\widetilde{M}_f]_{i,j} = d_f(x^{(i)}, \tilde{x}^{(j)})$, $x^{(i)}, \tilde{x}^{(j)} \sim P$, and maximizing EOT cost
  \begin{equation}\label{eq-sim-ex}
    W(P_f, Q_f) = \min_{{\pi} \in \Pi(P_f, Q_f) } \dotp{{\pi}}{M_f} - \la H({\pi}).
  \end{equation}
\item Learning of generator $g$ is minimizing EOT cost
  \begin{equation}
    W(P_f, Q_f).
  \end{equation}
\end{enumerate} 
The above alternating optimization is similar to triplet loss used in \cite{7298682}. Both $W(P_f, P_f)$ and $W(P_f, Q_f)$ have lower bounds, but no upper bounds. 
We combine the step 1 in above using a hinge loss and define the following costs.
\begin{equation}
  \hspace{-10pt}
  \begin{array}{rl}
    &\Ll_f(P_f, Q_f) \triangleq \max\left(0, W(P_f,P_f)- W(P_f,Q_f) + \gamma \right),  \\
    &\Ll_g(P_f,Q_f) \triangleq W(P_f,Q_f).
  \end{array}  
\end{equation}
where $\gamma>0$. Hinge loss helps to balance the adversarial
training of the $f$ and $g$.
Note the our hinge adversarial loss shares similarity only in form to the self-attention
GAN\cite{2018arXiv180508318Z} and geometric GAN\cite{2017arXiv170502894L}
but is motivated differently and defined in different metric. 
We used neural networks for constructing $f$ and $g$ functions. Let us assume that 
the parameters of $f$ and $g$ are $\omega$ and $\theta$, respectively.
Then the adversarial training between representation $f$ and
generator $g$ is the following alternative optimization problem:
\begin{equation}
  \begin{array}{rl}
    &\min_{f} \Ll_f(P_f, Q_f) = \min_{\omega} \Ll_f(P_f, Q_f) \\ 
    &\min_{g} \Ll_g(P_f, Q_f) = \min_{\th} \Ll_{g}(P_f, Q_f).
  \end{array}
\end{equation}
The EOTGAN is shown in Algorithm \ref{algo-eWGAN}.

\begin{algorithm}
  \caption{EOT based GAN (EOTGAN)}\label{algo-eWGAN}
  \begin{algorithmic}[1]
    \Require{$l$: the update rate at each iteration, $N$: the batch
      size and $\th_{0}$, $\omega_0$: the initial parameters for $g$ and $f$.}
    \While{$\theta$ has not converged}
    \State Sample two batches of data $ \left\{ x^{(i)}
    \right\}_{i=1}^{N}, \left\{ \tilde{x}^{(i)} \right\}_{i=1}^{N}  $,
    and latent samples $\left\{ z^{(i)} \right\}_{i=1}^{N} $,
    $x^{(i)},\tilde{x}^{(i)} \sim P, z\sim P_{z}$.
    \State Get $ \left\{ y^{(i)} \right\}_{i=1}^{N}$ by passing
    $\left\{ z^{(i)} \right\}_{i=1}^{N}$ through $g$.
    \State $\widetilde{\pi}^{\ast} \gets$ solving $ W_f\left( \left\{ f(x^{(i)})
      \right\}_{i=1}^{N} , \left\{ f(\tilde{x}^{(i)})
      \right\}_{i=1}^{N}\right)$
    \State ${\pi}^{\ast} \gets$ sovling $W_f\left( \left\{ f(x^{(i)})
      \right\}_{i=1}^{N}, \left\{ f({y}^{(i)})
      \right\}_{i=1}^{N} \right)$
    \State $\partial{f} \gets \nabla_{\om} \max\left(0,  \dotp{\widetilde{\pi}^{\ast}}{\widetilde{M}}-\dotp{{\pi^{\ast}}}{{M}}+ \gamma\right)$
    \State $\om \gets \om - l \cdot \partial{f}$
    \State Sample $\left\{ z^{(i)} \right\}_{i=1}^{N}$
    and get $ \left\{ y^{(i)} \right\}_{i=1}^{N}$ via $g$.
    \State $\pi^{\ast} \gets$ sovling $W_f\left( \left\{ f(x^{(i)})
      \right\}_{i=1}^{N}, \left\{ f({y}^{(i)})
      \right\}_{i=1}^{N} \right)$
    \State $\partial{g} \gets \nabla_{\th} \dotp{\pi^{\ast}}{M}$
    \State $\th \gets \th -l \cdot \partial{g}$
    \EndWhile
  \end{algorithmic}
\end{algorithm}

\subsubsection{Advantage of EOT against OT}

Usage of entropy regularization in EOT avoids the need for Kantorovich-Rubinstein duality
of OT, thus is free from Lipschitz constraint. In literature, several methods endeavor to
satisfy Lipschitz constraint, for example, projecting neural network
parameters into a space fulfilling Lipschitz constraint via weight
clipping \cite{2017arXiv170107875A}, spectrum
normalization \cite{2018arXiv180205957M}), or adding gradient 
penalty into GAN's cost function \cite{2017arXiv170400028G}. Projecting
approaches bring the problem of neural network capacity underuse and
limit its ability to learn complex mapping. Gradient penalty approach takes
gradients of each layer's weight parameters of a neural network into
GAN's cost, thus computation complexity grows fast as the neural
network goes deeper. 
EOT avoids the above mentioned problems and also has the benefit of a lower
computation complexity. With entropy-regularization and Sinkhorn
algorithm, the computation complexity scales as $\mathcal{O}(N^2)$ \cite{2013arXiv1306.0895C}. 
On the other hand, solving OT cost using interior-point methods has computational requirement as 
$\mathcal{O}(N^3\log{N})$. 
\section{Experimental Results}
We perform experiments to justify our arguments on loss choice and algorithms. We evaluate our
generative models on a toy synthetic dataset of Gaussian-mixture distribution and real image digit dataset MNIST. 
\begin{figure*}[!ht]
  \captionsetup[subfigure]{justification=centering}
  \centering
  \begin{subfigure}[b]{0.44\textwidth}
    \centering
    \includegraphics[width=1\linewidth]{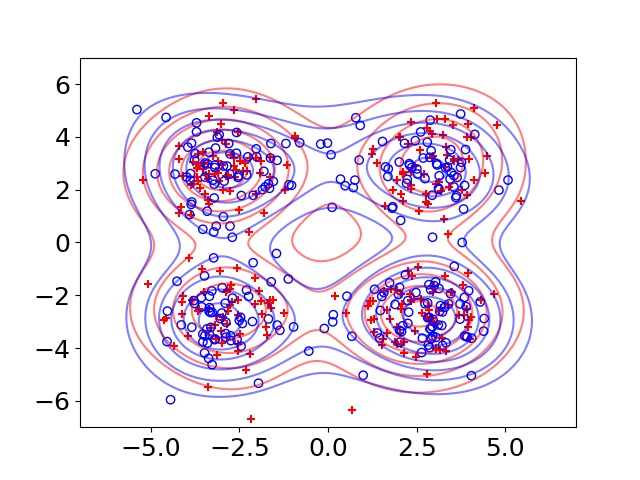}
    \caption{}
    \label{fig-toy}
  \end{subfigure}
  \centering
  \begin{subfigure}[b]{0.44\textwidth}
    \centering
 \includegraphics[width=0.8\linewidth]{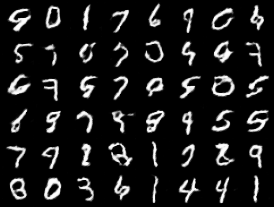}\vspace{8pt}
    \caption{}
    \label{fig-fake-wgan}
  \end{subfigure}
  \centering
  \begin{subfigure}[b]{0.44\textwidth}
    \centering
    \includegraphics[width=1.1\linewidth]{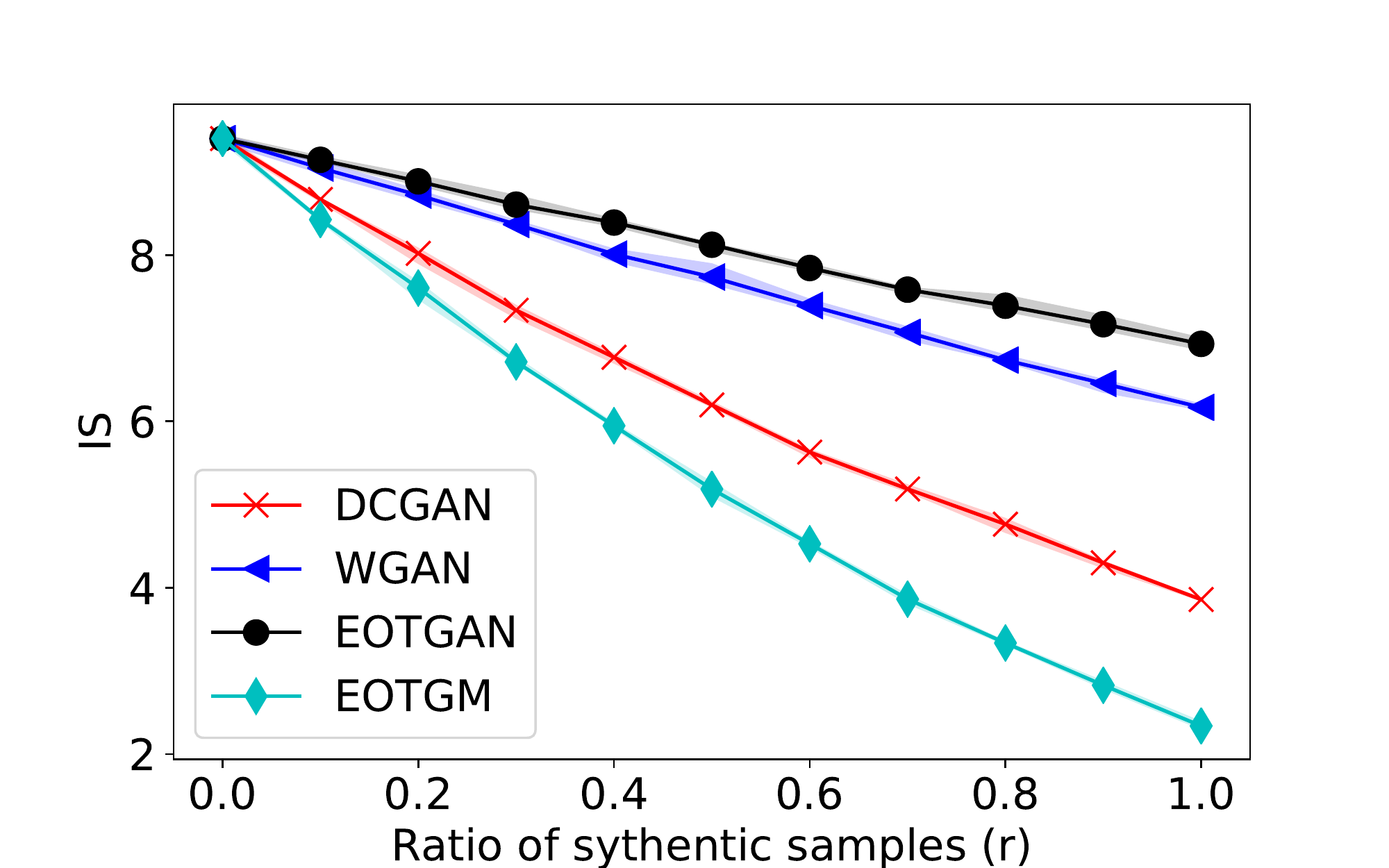}
    \caption{}
    \label{fig-tra-is}
  \end{subfigure}
  \begin{subfigure}[b]{0.44\textwidth}
    \centering
    \includegraphics[width=1.1\linewidth]{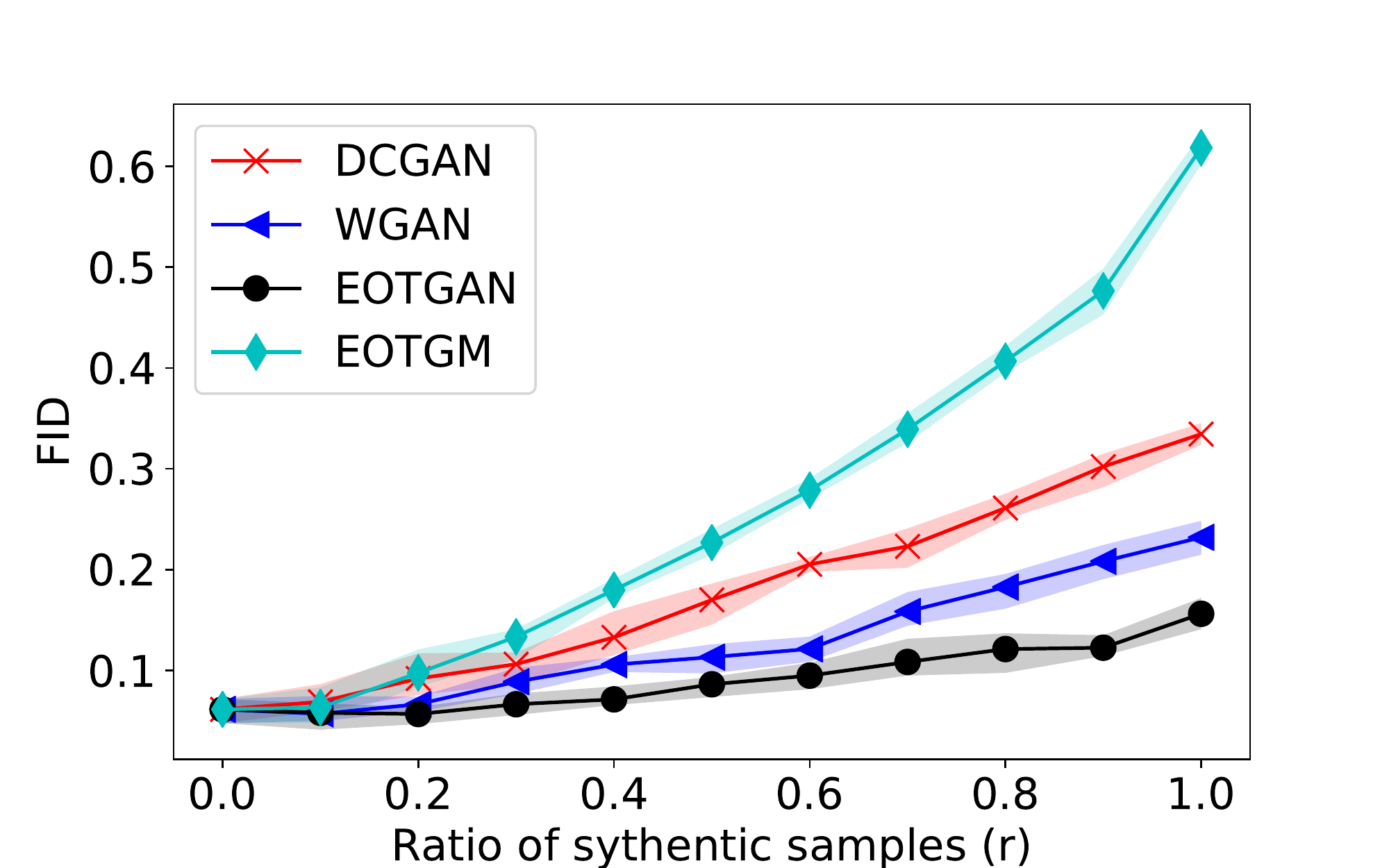}
    \caption{}
    \label{fig-tra-fid}
  \end{subfigure}
  \caption{(a) Toy distribution learning ($4$-mixture Gaussians) using EOTGM. Real samples (red '+') and contour
    (red curve), versus generated samples (blue 'o') and contour
    (blue curve) by $g$. (b) Generated samples by EOTGAN for MNIST dataset. (c) and (d)
    Comparison of IS and FID (on MNIST) versus mixing ratio $r$. (For
    each model at a certain mixture ratio, $5$ experiments are
    independently performed. Each solid curve with markers plots the mean of $5$
    experiments with shaded areas denoting the range of corresponding
    results.}
  \label{fig-tra-score}\vspace{10pt}  
\end{figure*}

\subsection{Evaluation Metrics}\label{subsec-metric}
Inception Score (IS) has been popularly used in evaluation of GAN
models\cite{NIPS2016_6125}. IS is defined as $
IS(Q) = \exp\bigl[ \EE_{y \sim Q} KL\bigl( \PP(c | y)\| \PP(c) \bigr)  \bigr]$,
where $x\sim Q$
indicates synthetic sample from distribution $Q$ induced by generator $g$, $KL(\cdot, \cdot)$ is Kullback-Leibler divergence, $\PP(c|y)$ is the
conditional class distribution, and $\PP(y) = \int_x\PP(c|y) \,d Q(x)$
is the marginal class distribution. Large IS score means generated
samples contain clear objects. Generative models with high IS can output high
diversity of samples.
Apart from $KL$-based metric,
an alternative common metric is Frechet Inception Distance (FID)\cite{2017arXiv170608500H}. FID measures the OT distance
of two probability distribution by assuming the two distributions are
Gaussian. Smaller FID means the generated samples are more similar to
empirical samples. Both FID and IS will be used in our experiments. High IS and low FID are better. 

\subsection{Evauation of EOTGM using toy dataset}\label{subsec-mg}

We firstly evaluate our proposed EOTGM on a toy dataset sampled from a known probability
distribution: two-dimensional four-mixture Gaussian. Here noise $P_z$ is Gaussian distributed:
$\mathcal{N}\left(\bigl(\begin{smallmatrix}& 0\\
    &0\end{smallmatrix}\bigr) ,\bigl( \begin{smallmatrix}1 & 0\\ 0 &
    1\end{smallmatrix}  \bigr)\right)$. The generator $g$ uses a neural network with structure: Dense $256$ $\rightarrow$ ReLU $\rightarrow$
Dense $256$ $\rightarrow$ ReLU $\rightarrow$ Dense $256$ $\rightarrow$
ReLU$\rightarrow$Dense $2$.
These empirical samples are used by
Algorithm~\autoref{algo-E-WL} (EOTGM) to train $g$.
The target distribution  for $g$ is the mixture
Gaussian. In Fig.\ref{fig-toy} we plot the empirical samples and the synthetic samples generated by $g$. The corresponding contours are
also plotted. It shows that the induced distribution by $g$ approaches the
mixture Gaussian distribution well without missing any mode.

\subsection{Evaluation of generative models using MNIST}
In this subsection we evaluate both the generative models using MNIST dataset.
The representation mapping $f$ in EOTGAN adapts two converlutional layers
appended with fully connected layers\footnote{$32$ Conv2d $5 \times5$
  $\rightarrow$ PReLU $\rightarrow$ MaxPool $2\times2$ $\rightarrow$
  $64$ Conv2d $5\times5$ $\rightarrow$ PReLU $\rightarrow$ MaxPool
  $2\times2$ $\rightarrow$ Dense $256$ $\rightarrow$ PReLU
  $\rightarrow$ Dense $256$ $\rightarrow$ PReLU $\rightarrow$ Dense
  $2$} similar to \cite{1467314}\cite{1640964}. Generator $g$ uses the same
setting as that of DCGAN and WGAN. Noise $P_z$ is $100$-dimensional Guassian.
We report IS and FID scores of EOTGAN in comparison with DCGAN and WGAN. Since EOTGAN is trained with representation mapping $f$ that
acts as feature mapping, it is not fair to use this representation mapping $f$ to do the
evaluation and make comparison since it would gives EOTGAN
advantages. Similar to \cite{2018arXiv180607755X}, we train a 34-layer
ResNet on MNIST to perform feature extraction for metric measurements
of IS and FID. In addition, we put EOTGM (Algorithm~\autoref{algo-E-WL}) in the comparison as well.

Data for evaluations is constructed by mixing empirical samples and
synthetic samples generated by $g$. We draw the set $\mathcal{S}_{\mathrm{em}}$ of 2000 empirical samples from MNIST dataset. To generate a set $\mathcal{S}_{\mathrm{syn}}$ of synthetic
samples we draw $2000r$ samples from the generator network $g$ where
$r\in[0,1]$ while rest $2000(1-r)$ are sampled directly from
MNIST. All following experiments are applied on
$\mathcal{S}_{\mathrm{em}}$ and $\mathcal{S}_{\mathrm{syn}}(r)$. 
The way of mixing empirical data and generated data helps
us to identify if a metric is intuitively helpful. Among the chosen metrics IS at $r=0$ serves as the upper bound for the test while the FID score at $r=0$ serves as the lower bounds for the corresponding tests. 

IS measures how certain a classifier assigns a class label to a given
generated sample. The larger IS is, the better the generative model
is. We plot IS versus $r$ for different models in
Fig. \ref{fig-tra-is}. IS scores of all four tested models drop with
increasing portion of synthetic samples in
$\mathcal{S}_{\mathrm{syn}}$, which is consistent with intuition. IS
of EOTGAN drops at the slowest rate among the four model as more
synthetic samples, for larger $r$, are mixed into test data. It shows
that EOTGAN outperforms WGAN and DCGAN in this test. EOTGM is found to provide the lowest IS.
This may be attributed to the setup that 
EOT optimization with cost measured by Euclidean distance of signals
fails to capture semantic similarity.

In Fig.~\ref{fig-tra-fid} the perfomances of different models are
compared using the FID metric. The smaller the FID of a generative model is, the more similar the
generated samples are to the empirical samples. EOTGAN is the least affected model among all the four, as the ratio $r$ increases, i.e. the generated samples by EOTGAN is more similar to the empirical
ones in the feature space regarding to FID. FID of WGAN is larger than that of EOTGAN. As
more generated samples are mixed the FIDs of DCGAN and EOTGM grow
even faster, which means the samples generated by these two models are
less similar to the empirical samples.

\section{Conclusion}
This work shows that entropy-regularized optimal transport cost is
useful to train neural network based generative models for learning implicit probability
distributions. With computationally efficient Sinkhorn algorithm,
learning of a probability distribution by a generative model can be
posed as an one-shot optimization problem. For further progress in
quality of generating samples, our experiments show that additional use of
representation mapping and alternative optimization based on
adversarial game produce better
semantic samples. 

\bibliographystyle{IEEEbib}

\bibliography{myref}


\end{document}